\documentclass{article}

    \PassOptionsToPackage{numbers, compress}{natbib}
 \usepackage[final]{neurips_2025_ml4ps} % added

\usepackage[utf8]{inputenc} % allow utf-8 input
\usepackage[T1]{fontenc}    % use 8-bit T1 fonts
\usepackage{hyperref}       % hyperlinks
\usepackage{url}            % simple URL typesetting
\usepackage{booktabs}       % professional-quality tables
\usepackage{amsfonts}       % blackboard math symbols
\usepackage{nicefrac}       % compact symbols for 1/2, etc.
\usepackage{microtype}      % microtypography
\usepackage{xcolor}         % colors

\usepackage{graphicx} % added

\title{Surrogate Neural Architecture Codesign Package (SNAC-Pack)}

\author{%
  Jason Weitz\\
  University of California San Diego\\
  La Jolla, CA 92093, USA \\
  \texttt{jdweitz@ucsd.edu} \\
  \And
  Dmitri Demler \\
  University of California San Diego\\
  La Jolla, CA 92093, USA\\
  \texttt{ddemler@ucsd.edu} \\
  \AND
  Benjamin Hawks \\
  Fermi National Accelerator Laboratory \\
  Batavia, IL 60510, USA \\
  \texttt{bhawks@fnal.gov} \\
  \And
  Nhan Tran \\
  Fermi National Accelerator Laboratory \\
  Batavia, IL 60510, USA \\
  \texttt{ntran@fnal.gov} \\
  \And
  Javier Duarte \\
  University of California San Diego\\
  La Jolla, CA 92093, USA \\
  \texttt{jduarte@ucsd.edu} \\
}

\begin{document}

\begin{flushright}
FERMILAB-CONF-25-0834-CSAID
\end{flushright}

\maketitle

\begin{abstract}

Neural Architecture Search is a powerful approach for automating model design, but existing methods struggle to accurately optimize for real hardware performance, often relying on proxy metrics such as bit operations.
We present Surrogate Neural Architecture Codesign Package (SNAC-Pack), an integrated framework that automates the discovery and optimization of neural networks focusing on FPGA deployment.
SNAC-Pack combines Neural Architecture Codesign's multi-stage search capabilities with the Resource Utilization and Latency Estimator, enabling multi-objective optimization across accuracy, FPGA resource utilization, and latency without requiring time-intensive synthesis for each candidate model. 
We demonstrate SNAC-Pack on a high energy physics jet classification task, achieving 63.84\% accuracy with resource estimation.
When synthesized on a Xilinx Virtex UltraScale+ VU13P FPGA, the SNAC-Pack model matches baseline accuracy while maintaining comparable resource utilization to models optimized using traditional BOPs metrics.
This work demonstrates the potential of hardware-aware neural architecture search for resource-constrained deployments and provides an open-source framework for automating the design of efficient FPGA-accelerated models.

\end{abstract}

\section{Introduction}

% \begin{enumerate}
%     \item framework; modular - choose own metrics, surrogate model
%     \item build on existing tools
%     \item this is the new contribution
% \end{enumerate}

Machine learning models are applied across diverse domains, achieving state-of-the-art performance.
Yet, deploying these models in resource-constrained environments, such as edge devices, is challenging due to strict hardware and latency requirements.
In order to meet these constraints, researchers have developed methods such as automated architecture design and model compression.
Neural Architecture Search (NAS) can automate model design, but existing approaches typically optimize for accuracy alone or use simplistic computational metrics like bit operations (BOPs) that approximate actual hardware performance. 

To address this challenge, we introduce Surrogate Neural Architecture Codesign Package (SNAC-Pack), an integrated software framework that automates the discovery and optimization of neural network architectures specifically tailored for hardware deployment.
SNAC-Pack combines two tools: Neural Architecture Codesign~\cite{Weitz_2025}, which performs a multi-stage neural architecture search to discover optimal models, and the Resource Utilization and Latency Estimator~\cite{Rahimifar_2025}, which uses a surrogate model to predict how an architecture will perform when synthesized for implementation on an FPGA.

By integrating a fast and accurate surrogate model for resource estimation directly into the search stage, SNAC-Pack can perform a more effective multi-objective optimization for accuracy, hardware resource usage, and latency while avoiding time-intensive full hardware synthesis for every candidate model.
This work explores the use of SNAC-Pack on a jet classification task~\cite{pierini_2020_3602260}, from a search space to a synthesized model. The SNAC-Pack software is available \href{https://github.com/fastmachinelearning/nac-opt.git}{here}.

%%%%
% Machine learning has become a critical tool for analysis and decision-making across a wide range of scientific domains, from particle physics to materials science. However, the deployment of neural networks in resource-constrained environments, such as hardware accelerators and edge devices, remains a significant challenge. This often requires specialized expertise in both neural architecture design and hardware optimization. 

% To address this challenge, we introduce the Surrogate Neural Architecture Codesign Package (SNAC-Pack), an integrated framework that automates the discovery and optimization of neural network architectures specifically tailored for hardware deployment. SNAC-Pack combines two tools: Neural Architecture Codesign \cite{Weitz_2025}, which performs a multi-stage neural architecture search to discover optimal models, and the Resource Utilization and Latency Estimator \cite{Rahimifar_2025}, which predicts how an architecture will perform when synthesized for implementation on an FPGA.

% SNAC-Pack streamlines the neural architecture design process by enabling researchers to automatically explore diverse architectures optimized for both task performance and hardware efficiency. By providing quick estimates of resource utilization and latency without requiring time consuming synthesis, SNAC-Pack accelerates the development cycle. State-of-the-art compression techniques, such as quantization-aware training and pruning, further optimize the models, resulting in architectures that can be deployed to FPGA hardware.
%%%%

\section{Related Work}

% \begin{enumerate}
%     \item nac
%     \item rule4ml
%     \item this is the new contribution
% \end{enumerate}
Neural Architecture Search (NAS)~\cite{9508774,zoph2016neural} is an automated process that explores a search space of model architectures with a search strategy based on a set of evaluation objectives, such as accuracy. 
Neural Architecture Codesign (NAC)~\cite{Weitz_2025} is a multi-stage NAS, consisting of a global and local search. The global search explores a wide range of architectures, resulting in a Pareto front of well-performing models. Selecting an architecture along this front, local search performs model compression, including quantization-aware-training (QAT)~\cite{quantization_survey} and pruning~\cite{modelcompression}. With this further refinement, the model is then synthesized with hls4ml~\cite{hls4ml,fastml_hls4ml}.

Separately, surrogate models~\cite{10.1145/3489517.3530408,10.1145/3240765.3264635} have been developed to accelerate the design cycle by providing rapid feedback on hardware performance without time-consuming synthesis runs. The rule4ml library~\cite{Rahimifar_2025}, for instance, introduces a method to accurately predict the resource utilization, including block ram (BRAM), digital signal processors (DSPs), flip flops (FFs), lookup tables (LUTs), initiation interval (II), and latency of a neural network on an FPGA. 

\section{Method}

The SNAC-Pack tool builds upon NAC by introducing the additional objectives that can be estimated with rule4ml.
With NAC integration, the global search stage begins with a user-defined search space that specifies the range of possible architectures, such as the types of layers, number of neurons, activations, and other hyperparameters.
A multi-objective search algorithm then explores this space, samples candidate architectures, and performs evaluation. 
Any combination of the metrics estimated by rule4ml, BOPs, and accuracy can be used as objectives in the search.
In this work, SNAC-Pack utilizes an average of the resource utilization estimation, average clock cycles, and accuracy.
With a Pareto front produced by global search, an optimal architecture is selected based on the user's specific accuracy and resource constraints. This model then proceeds to the local search stage, which focuses on further refinement with QAT and iterative magnitude pruning. Another Pareto front is produced from which a model with a specific bit precision and sparsity.
Finally, the fully optimized model is synthesized using the hls4ml library to generate high-level synthesis (HLS) code for FPGA deployment. 

% \begin{enumerate}
%     \item utilizing nac pipeline for process
%     \item new: implementing rule4ml for new resource-aware evaluation to improve codesign
%     \item explain this addition within the process for hardware and latency constrained tasks. Ability to perform multi-objective optimization with accuracy, resources, and clock cycles as objectives.
% \end{enumerate}

\section{Jet Classification Implementation}

To show the effectiveness of SNAC-Pack, we apply it to jet classification, a common and challenging task in high energy physics at the Large Hadron Collider (LHC).
The goal is to accurately classify collision-created jets into one of five categories (light quark, gluon, W boson, Z boson, top quark) based on their kinematic properties.
This is showcased with the hls4ml LHC dataset \cite{pierini_2020_3602260}.
The 8 constituents with the greatest transverse momentum are used per jet.

For this task, we configure SNAC-Pack to search for an optimal multi-layer perceptron (MLP) architecture, with an 8 constituent MLP as a comparative baseline \cite{Odagiu_2024} with the data processed and normalized as done there.
This baseline is chosen, as it is one of the state-of-the-art architectures for this task.
The global search uses the Non-dominated Sorting Generic Algorithm II (NSGA-II) \cite{NSGA}, exploring a search space of varying number of layers, hidden units per layer, activations, and batch normalization \cite{ioffe2015batch}, as seen in Table \ref{tab:mlp-parameter-table}. 
The objectives used are estimated average hardware utilization, estimated clock cycles, and accuracy. All training is performed with a batch size of 128. Global search is performed for 500 trials, 5 epochs per trial, and an evolutionary population size of 20.

\begin{table}[t]
  \caption{Comprehensive parameter space for the multilayer perceptron (MLP) search.}
  \label{tab:mlp-parameter-table}
  \centering
  \small
  \renewcommand{\arraystretch}{1.2}
  \begin{tabular}{ll}
    \toprule
    \textbf{Parameter} & \textbf{Space} \\
    \midrule
    Number of layers        & \{4, 5, 6, 7, 8\} \\
    Hidden units per layer  & \\
    \quad Layer 1 & \{64, 120, 128\} \\
    \quad Layer 2 & \{32, 60, 64\} \\
    \quad Layer 3 & \{16, 32\} \\
    \quad Layer 4 & \{32, 64\} \\
    \quad Layer 5 & \{32, 64\} \\
    \quad Layer 6 & \{32, 64\} \\
    \quad Layer 7 & \{16, 32\} \\
    \quad Layer 8 & \{32, 44, 64\} \\
    Activation function     & \{ReLU, Tanh, Sigmoid\} \\
    Batch normalization     & \{True, False\} \\
    Learning rate           & \{0.0010, 0.0015, 0.0020\} \\
    L1 regularization       & \{0.0, $10^{-6}$, $10^{-5}$, $10^{-4}$\} \\
    Dropout rate            & \{0.0, 0.05, 0.1\} \\
    \bottomrule
  \end{tabular}
\end{table}

\begin{figure}[t]
  \centering
  \includegraphics[width=0.6\linewidth]{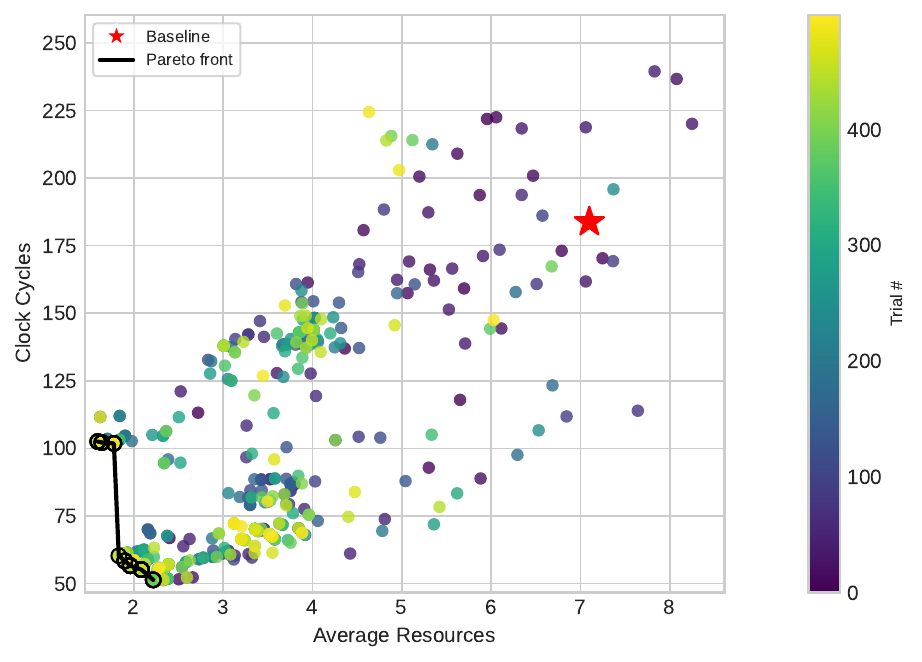} % larger
  \caption{SNAC-Pack Pareto front of estimated average resources versus estimated clock cycles. Each point represents a unique architecture sampled.}
  \label{fig:snac-pack_res_cc}
\end{figure}

\begin{figure}[t]
  \centering
  \includegraphics[width=0.6\linewidth]{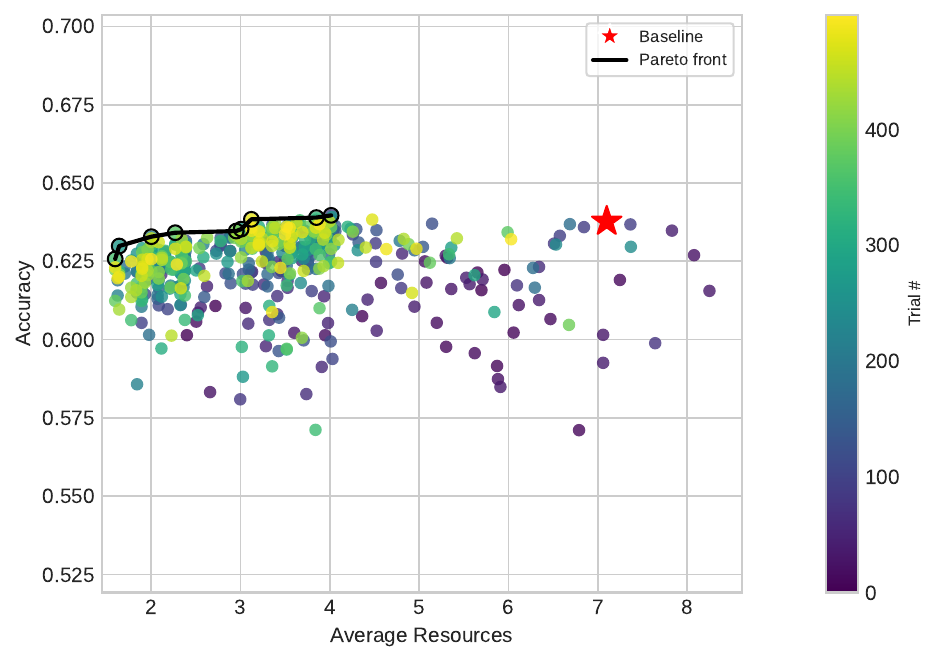} % larger
  \caption{SNAC-Pack Pareto fronts of estimated average resources versus accuracy. Each point represents a unique architecture sampled.}
  \label{fig:snac-pack_res_acc}
\end{figure}

\begin{figure}[t]
  \centering
  \includegraphics[width=0.6\linewidth]{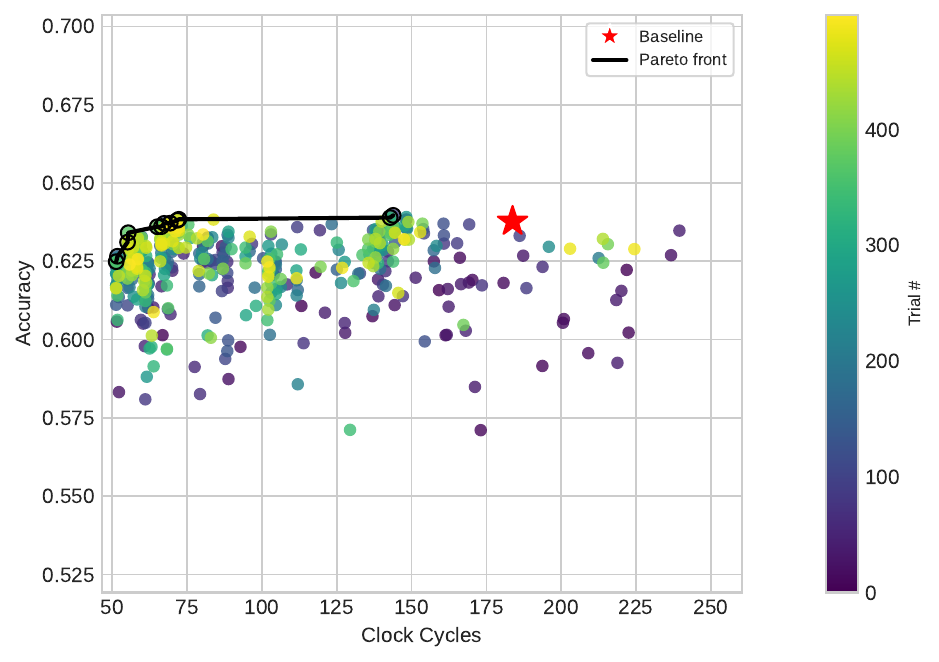} % larger
  \caption{SNAC-Pack Pareto fronts of estimated clock cycles versus accuracy. Each point represents a unique architecture sampled.}
  \label{fig:snac-pack_cc_acc}
\end{figure}

\begin{figure}[t]
  \centering
  \includegraphics[width=0.6\linewidth]{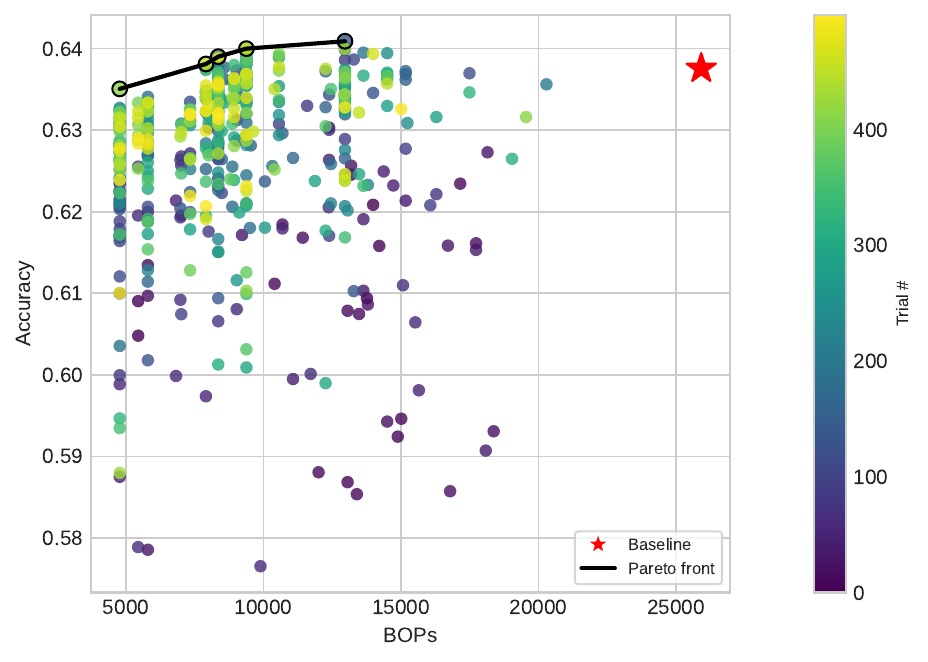} % smaller
  \caption{NAC Pareto front of BOPs versus accuracy. Each point represents a unique architecture sampled.}
  \label{fig:nac_pareto_front}
\end{figure}

The result is depicted as three Pareto fronts, seen in Figs. \ref{fig:snac-pack_res_cc}, \ref{fig:snac-pack_res_acc}, \ref{fig:snac-pack_cc_acc}. 
For comparison the search is also ran with NAC, optimizing solely for BOPs and accuracy, with the same number of trials and epochs.
The resulting Pareto front is shown in Fig.~\ref{fig:nac_pareto_front}.
The Pareto-optimal architectures with an accuracy greater than 0.638 are selected for local search and compared to the baseline~\cite{Odagiu_2024}, seen in Table~\ref{tab:mlp-global-comparison}. This accuracy value is selected as the threshold to ensure the accuracy meets or exceeds that of the baseline.

For local search, each selected architecture has a 5 epoch warm-up, followed by 10 iterations of iterative magnitude pruning~\cite{frankle2019stabilizing,frankle2018lottery}, each 10 epochs, with 20\% pruned per iteration, with QAT at 8-bit precision.

With hls4ml, the resulting architectures are then synthesized on the Xilinx Virtex UltraScale+ VU13P FPGA, with io\_parallel io\_type, latency strategy, a reuse factor of 1. Resource utilization and latency is shown in Table \ref{tab:model-comparison-synth}.

Seen in Table \ref{tab:mlp-global-comparison}, the model found with the SNAC-Pack framework performs similarly to the NAC model and baseline in terms of accuracy, and it is comparable to the NAC model in the other criteria. Table \ref{tab:model-comparison-synth} shows that the SNAC-Pack model matches metrics of the other models, but can improve in terms of latency. This is an indicator of a need to improve the estimation of resources themselves.

\begin{table}[t]
  \caption{Comparison of model accuracy, BOPs, and estimated hardware metrics from global search. Note that while all metrics are reported here for consistency, the Baseline was optimized for accuracy, NAC for accuracy and BOPs, and SNAC-Pack for accuracy, estimated average resources and clock cycles. The best values are reported in bold.}
  \label{tab:mlp-global-comparison}
  \centering
  \footnotesize
  \begin{tabular}{lcccc}
    \toprule
    Model & Accuracy [\%] & BOPs & Est. average resources & Est. clock cycles \\
    \midrule
    Baseline \cite{Odagiu_2024}          & 63.77 & 25{,}916 & 7.10 & 183.74 \\
    Optimal NAC \cite{Weitz_2025}      & 63.81 & \textbf{7{,}904}  & 3.60   & \textbf{62.69} \\
        Optimal SNAC-Pack & \textbf{63.84} & 8{,}352  & \textbf{3.12} & 72.24 \\
    \bottomrule
  \end{tabular}
\end{table}

% \begin{figure}[t]
%   \centering
%   \includegraphics[width=0.75\linewidth]{figures/pruning_curves.pdf} % larger
%   \caption{Pareto fronts of validation accuracy versus estimated hardware resources and clock cycles.}
%   \label{fig:snac-pack_pareto_fronts}
% \end{figure}

% \begin{table}[t]
%   \caption{Hardware resource utilization and latency estimates for the selected models. cc is the number of clock cycles. this is with mixed boards (bottom 2 are u-200 synth}
%   \label{tab:model-comparison-synth}
%   \centering
%   \scriptsize
%   \resizebox{\textwidth}{!}{%
%     \begin{tabular}{lcccccc}
%       \toprule
%       Model & Lat. [ns] (cc) & II [ns] (cc) & DSP & LUT & FF & BRAM \\
%       \midrule
%       Baseline \cite{Odagiu_2024} & 105 (21) & 5 (1) & 262 (2.1\%) & 155080 (9.0\%) & 25714 (0.7\%) & 4 (0.1\%) \\
%       Optimal NAC \cite{Weitz_2025} & 130 (26) & 5 (1) & 0 (0\%) & 53712 (4.54\%) & 12122 (0.51\%) & 8 (0.37\%) \\
%       Optimal SNAC-Pack & 139 (24) & 5 (1) & 0 (0\%) & 56921 (4.81\%) & 12693 (0.53\%) & 0 (0\%) \\
%       \bottomrule
%     \end{tabular}%
%   }
% \end{table}

\begin{table}[t]
  \caption{Hardware resource utilization and latency estimates for the selected models. The baseline is pruned by 50\% and quantized to 8 bits. NAC and SNAC-Pack models are both synthesized after their respective local searches. CC is the number of clock cycles. The selected architectures are the result of pruning to approximately 50\% and quantization aware training at 8 bit precision. The best values are reported in bold.}
  \label{tab:model-comparison-synth}
  \centering
  \scriptsize
  \resizebox{\textwidth}{!}{%
    \begin{tabular}{lcccccc}
      \toprule
      Model & Lat. [ns] (cc) & II [ns] (cc) & DSP & LUT & FF & BRAM \\
      \midrule
      Baseline \cite{Odagiu_2024} & \textbf{105 (21)} & \textbf{5} (1) & 262 (2.1\%) & 155080 (9.0\%) & 25714 (0.7\%) & 4 (0.1\%) \\
      Optimal NAC \cite{Weitz_2025} & 125 (25) & 60 (12) & \textbf{0} & \textbf{54075 (3.13\%)} & \textbf{12016 (0.35\%)} & 8 (0.3\%) \\
      Optimal SNAC-Pack & 140 (24) & 70 (12) & \textbf{0} & 57728 (3.34\%) & 12605 (0.36\%) & \textbf{0} \\
      \bottomrule
    \end{tabular}%
  }
\end{table}

\section{Conclusion}

% \begin{enumerate}
%     \item Resource estimation compared to original ultrafast implementation (difficult to compare to nac directly, as that only used deep sets). Ran nac (acc and BOPs) to compare
%     \item importance of hardware-aware searches for resource and latency constrained tasks
%     \item future work with wa-hls4ml dataset and surrogate models \cite{10.1145/3706628.3708827}
%     \item Emphasize future work: snac-pack ideally finds a better architecture faster - that is that it finds architectures that are lower in latency and use fewer resources more effectively than NAC's BOP optimization method.
% \end{enumerate}

This work introduced the Surrogate Neural Architecture Codesign Package (SNAC-Pack), a framework that extends NAC by incorporating resource-aware objectives estimated with rule4ml.
Applied to the jet classification task, the optimal model produced by SNAC-Pack performed comparably to the NAC method and baseline reference.

With slight under performance, this new framework establishes the potential of hardware-awareness in NAS.
As a functioning pipeline, there is an opportunity for extensions. 
Incorporating additional surrogate models trained on large datasets \cite{hawks2025wahls4mlbenchmarksurrogatemodels} in future work can be an enhancement to improve SNAC-Pack, realtive to the BOPs proxy.
With this, the estimation of resources can be refined in order to discover lower latency architectures that utilize fewer true resources.

\clearpage
\section{Acknowledgments}
This manuscript has been authored by Fermi Forward Discovery Group, LLC under Contract No. 89243024CSC000002 with the U.S. Department of Energy, Office of Science, Office of High Energy Physics.
BH and NT are supported by Fermi Research Alliance, LLC under Contract No. DE-AC02-07CH11359 with the United States Department of Energy (DOE), Office of Science, Office of High Energy Physics.
BH and NT are also supported under the DOE Early Career Research program under Award No. DE-0000247070.
BH, JD, and NT are supported by the U.S. Department of Energy (DOE), Office of Science, Office of Advanced Scientific Computing Research under the ``Real-time Data Reduction Codesign at the Extreme Edge for Science'' Project(DE-FOA-0002501).
JD is also supported by the DOE, Office of Science, Office of High Energy Physics Early Career Research program under Grant No. DE-SC0021187, and the U.S. National Science Foundation (NSF) Harnessing the Data Revolution (HDR) Institute for Accelerating AI Algorithms for Data Driven Discovery (A3D3) under Cooperative Agreement No. PHY-2117997.
JW is supported by a WATCHEP fellowship sponsored by the DOE, Office of High-Energy Physics under Award No. DE-SC-0023527.

% \begin{enumerate}
%     \item nac
%     \item rule4ml
%     \item ultrafast jet classification
%     \item wa-hls4ml as future work implementation
% \end{enumerate}
\clearpage
\bibliographystyle{unsrtnat}
\bibliography{references}

\end{document}